\documentclass[11pt]{article}

\usepackage[preprint]{acl}

\usepackage{times}
\usepackage{latexsym}
\usepackage{comment} 
\usepackage[font=small]{caption}
\usepackage[T1]{fontenc}

\usepackage[utf8]{inputenc}

\usepackage{microtype}

\usepackage{inconsolata}

\usepackage{graphicx}
\usepackage{booktabs}
\usepackage{color}
\usepackage{tabularray}
\usepackage[most]{tcolorbox}
\usepackage{latexsym}
\usepackage{stmaryrd}
\usepackage{pifont}
\usepackage{hyperref}
\usepackage{xcolor}
\usepackage{float}
\usepackage{pifont}
\usepackage{linguex}

\usepackage[most]{tcolorbox}
\usepackage{lipsum}
\usepackage{graphicx}

\tcbset{
  chatbubble/.style={
    boxrule=0pt,
    colback=#1!20,
    colframe=#1!50,
    enhanced,
    left=6pt,right=6pt,top=6pt,bottom=6pt,
    width=0.9\columnwidth,
  }
}

\definecolor{Grandis}{rgb}{1,0.819,0.49}
\definecolor{Zanah}{rgb}{0.792,0.905,0.792}
\definecolor{TitanWhite}{rgb}{0.894,0.894,1}
\definecolor{Thistle}{rgb}{0.866,0.741,0.866}
\definecolor{SnowyMint}{rgb}{0.792,1,0.792}
\definecolor{Pippin}{rgb}{1,0.894,0.894}

\tcbset{
  mypromptbox/.style={
    enhanced,
    colback=red!5!white,
    colframe=red!25!white,
    coltitle=black,
    fonttitle=\bfseries,
    boxrule=0.5pt,
    arc=3pt,
    outer arc=3pt,
    left=6pt,
    right=6pt,
    top=6pt,
    bottom=6pt,
    boxsep=4pt,
    title={The prompt for \textit{UND classifier}}
  }
}

\tcbset{
  mypromptbox2/.style={
    enhanced,
    colback=green!5!white,
    colframe=green!25!white,
    coltitle=black,
    fonttitle=\bfseries,
    boxrule=0.5pt,
    arc=3pt,
    outer arc=3pt,
    left=6pt,
    right=6pt,
    top=6pt,
    bottom=6pt,
    boxsep=4pt,
    title={The prompt for \textit{Rewriter models}}
  }
}

\tcbset{
  mypromptbox16/.style={
    enhanced,
    colback=blue!5!white,
    colframe=blue!25!white,
    coltitle=black,
    fonttitle=\bfseries,
    boxrule=0.5pt,
    arc=3pt,
    outer arc=3pt,
    left=6pt,
    right=6pt,
    top=6pt,
    bottom=6pt,
    boxsep=4pt,
    title={\textbf{The Prompt for \textit{QA models}}}
  }
}

\title{\textit{Who is the richest club in the championship?} Detecting and Rewriting Underspecified Questions Improve QA Performance}

\author{Yunchong Huang \\
  ILLC, University of Amsterdam \\
  \texttt{franzhuang027@gmail.com} \\\And
  Gianni Barlacchi \\
  Amazon AGI \thanks{Work done outside Amazon.}\\
  \texttt{gbarlac@amazon.com} \\\AND 
  Sandro Pezzelle \\
  ILLC, University of Amsterdam \\
  \texttt{s.pezzelle@uva.nl}}

\begin{document}
\maketitle
\begin{abstract}
Large language models (LLMs) perform well on well-posed questions, yet standard question-answering (QA) benchmarks remain far from solved. We argue that this gap is partly due to \textit{underspecified questions}—queries whose interpretation cannot be uniquely determined without additional context. To test this hypothesis, we introduce an LLM-based classifier to identify underspecified questions and apply it to several widely used QA datasets, finding that 16\% to over 50\% of benchmark questions are underspecified and that LLMs perform significantly worse on them. To isolate the effect of underspecification, we conduct a controlled rewriting experiment that serves as an upper-bound analysis, rewriting underspecified questions into fully specified variants while holding gold answers fixed. QA performance consistently improves under this setting, indicating that many apparent QA failures stem from question underspecification rather than model limitations. Our findings highlight underspecification as an important confound in QA evaluation and motivate greater attention to question clarity in benchmark design.
\end{abstract}




\section{Introduction} 
Large Language Models (LLMs) perform well on clearly specified factual queries, yet widely used question-answering (QA) benchmarks remain unsolved \citep{Kwiatkowski2019,yang-etal-2018-hotpotqa,joshi-etal-2017-triviaqa,krishna-etal-2025-fact, sorodoc-etal-2025-garage}. In parallel, a growing body of work shows that LLMs struggle with vague, ambiguous, or incomplete queries \citep{Tanjim2025,liu-etal-2023-afraid, zhang-etal-2024-clamber,zhang-choi-2025-clarify, Tanjim2025bis, qian-etal-2024-tell, Herlihy2024OnOM}.

In this paper, we argue that these two observations are closely related. Specifically, we hypothesize that a non-trivial portion of the apparent difficulty of QA benchmarks stems not from model limitations alone, but from the presence of underspecified questions, that is, queries whose interpretation cannot be uniquely determined without additional contextual information, which is unavailable at evaluation time.


\begin{figure}[t]
    \centering
    \includegraphics[width=0.9\linewidth]{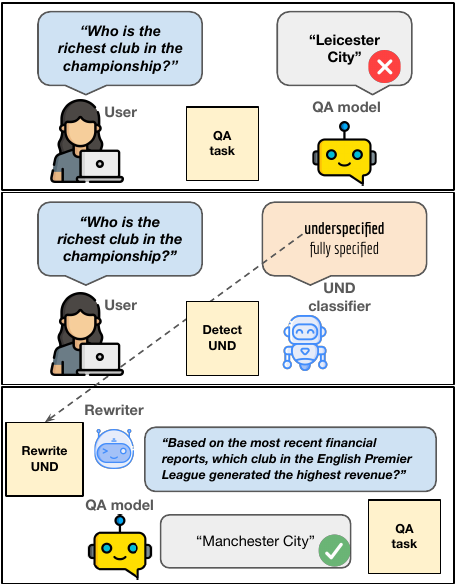}
\caption{Top: One real question from Natural Questions (NQ) dataset \cite{Kwiatkowski2019} and corresponding \textit{wrong} answer by a QA model, i.e., \texttt{GPT-4o} \cite{openai2024gpt4ocard}. Middle: We build an LLM-based classifier to detect underspecified (UND) questions in QA benchmarks. Bottom: We turn UND questions into fully specified ones using an LLM-based rewriter and verify that, by removing underspecification, performance on the QA task significantly improves.}
    \label{fig: Overall Workflow}
\end{figure}

The notion of underspecification originates in linguistic theory, where it describes expressions that admit multiple possible meanings and require additional contextual information to be resolved \citep{Egg2010, sep-ambiguity, vanBerkum2008,sep-vagueness, vanRooij2011, Kennedy2011,Carston2002, Belleri2014-BELSUA,Grice1957-GRIM}.
An underspecified utterance does not fully encode its intended meaning and relies on shared common ground, discourse context, or pragmatic inference to narrow down its interpretation \citep{Frisson2009, Harris2020b, pezzelle-2023-dealing, wildenburg-etal-2024-pre, Bach2004-BACCEM, Stalnaker2002-STACG}. While such mechanisms are often available in natural communication, in the context of QA benchmarks such information is typically unavailable, making it impossible to recover the intended answer from the question alone uniquely.
The top panel of Fig.~\ref{fig: Overall Workflow} illustrates this issue with an example from the Natural Questions (NQ) dataset \cite{Kwiatkowski2019}---\textit{Who is the richest club in the championship?}---where the specific referent of \textit{championship}, the definition of \textit{richest}, and the specific time frame to answer the question are all left underspecified. 

We hypothesize that underspecified questions are prevalent in existing QA benchmarks and systematically reduce reported model performance. To test this hypothesis, we first introduce an LLM-based classifier that distinguishes between fully specified (FS) and underspecified (UND) questions. Applying it to several widely used QA datasets, we find that all benchmarks contain a substantial proportion of underspecified questions, ranging from around 16\% of total questions in TriviaQA \citep{joshi-etal-2017-triviaqa} to over 50\% in FRAMES \citep{krishna-etal-2025-fact}. We then show that state-of-the-art LLMs perform significantly worse on UND questions than on FS ones across datasets. To further isolate the effect of underspecification, we conduct a controlled rewriting experiment that serves as an upper-bound analysis. By rewriting underspecified questions into fully specified variants while holding the gold answer fixed, we evaluate how QA models perform when underspecification is removed, and find that QA performance consistently improves on rewritten questions, indicating that many errors stem from question formulation rather than model limitations.

Our contributions are as follows: (1) we provide a systematic analysis of underspecification in widely used QA benchmarks, showing that it is both prevalent and impactful to evaluation reliability; (2) we introduce an automated method for identifying underspecified questions; (3) we demonstrate, via controlled rewriting experiments, that many apparent QA failures disappear once underspecification is removed; and (4) we release our data and code publicly to support reproducibility and future research at \url{https://github.com/franzyellow/Underspecification-QA-conf-paper}.

\section{Approach}



\paragraph{Step 1: Detecting underspecified questions}



We first develop an LLM-based classifier (the \textit{UND classifier}) to distinguish fully specified (FS) from underspecified (UND) queries, and use it to extract FS and UND subsets from standard QA benchmarks (see middle panel in Fig.~\ref{fig: Overall Workflow}). To ensure reliability, we select the best-performing model from a suite of SotA open-weight LLMs (e.g., Qwen3, DeepSeek R1)  by evaluating them on existing underspecification-annotated data. We further strengthen this classification by validating the model against a smaller, expert-annotated gold standard subset.

\paragraph{Step 2: Assessing LLM performance on QA}
With the benchmarks partitioned, we evaluate two state-of-the-art proprietary LLMs (the \textit{QA models}) on four diverse QA benchmarks. This step is a comparative analysis designed to test our core hypothesis: that standard accuracy metrics are systematically diminished by the UND portion of the data.

\paragraph{Step 3: Rewriting UND questions}
To confirm that lower performance on UND questions is due to missing context rather than a lack of model knowledge, we conduct a controlled rewriting experiment. We use an LLM-based rewriter to resolve the ambiguities identified in Step 1. Critically, this rewriter has access to the ground-truth (gold) answers and the classifier’s reasoning. This ``oracle'' setup serves as a controlled intervention to transform UND questions into FS variants while holding the intended answer fixed. 



\paragraph{Step 4: Reassessing LLM's QA performance}
In this final step, we reassess the performance of the \textit{QA models} from Step 2 on the rewritten UND questions and compare it to their performance on the original UND queries, evaluating whether rewriting leads to significant improvements. This step establishes an upper bound of performance, demonstrating how much of the failure disappears when the benchmark's underspecified elements are removed. 

\section{Experiments}

\subsection{Step 1: Detecting underspecified questions}

\paragraph{Data} To build a reliable \textit{UND classifier}, we curated UNDER, a multi-source dataset with FS/UND labels and comprising 855 questions from CLAMBER \citep{zhang-etal-2024-clamber}, IN3 \citep{qian-etal-2024-tell}, and CoCoNot \citep{brahman2024}.\footnote{
We did not consider the popular AmbigNQ dataset \cite{min-etal-2020-ambigqa} for two reasons: (i) its bottom-up crowdsourcing procedure introduced substantial cross-annotation inconsistencies on the perceptive threshold of underspecification; (ii) annotators were instructed to enumerate all plausible answers using Wikipedia, resulting in a much finer-grained and knowledge-intensive notion of underspecification. Thus, automatically mapping AmbigNQ instances to binary UND/FS labels without heavy manual verification is unreliable and, therefore, incompatible with the procedure used for the other datasets. However, we did consider AmbigNQ to construct a small, hand-curated ``gold'' test set, as explained below.} UNDER contains 855 questions in total, i.e., 431 FS (50.4\%), and 424 UND questions (49.6\%) encompassing different types of underspecification. These questions are annotated with different terminologies in corresponding source datasets, but have been mapped to the unified UND/FS binary labeling in our configuration.

To ensure linguistic rigor, we further developed UNDER-gold, a subset of 150 instances from the same datasets and from uncontroversially annotated examples (verified by the authors) in AmbigNQ \cite{min-etal-2020-ambigqa}, with no overlap with UNDER. Each QA pair was manually reviewed by an author with expertise in formal linguistics to verify the alignment between automated labels and theoretical underspecification. The manual annotation followed our working taxonomy of underspecification showcased in Table \ref{tab: taxonomy}\footnote{See Appendix~\ref{app:taxonomy} for a full illustration of the taxonomy.}, ensuring that each UND/FS label is grounded in principled reasoning and expert-verified and that all UND questions are identified with a finer-grained taxonomical type.

\begin{table}[ht]
\captionsetup{width=0.4\textwidth}
\centering
\scriptsize
\begin{tblr}{
  width = \linewidth,
  colspec = {Q[321]Q[467]Q[152]},
  row{1} = {Grandis},
  hlines,
  vlines,
}
UND types                                       & Example                                                                                                                           & Source  \\
\textbf{Type 1}: Missing necessary components            & {\textit{Ok Google, what's the capital?}\\\textcolor{red}{Which country/region is being asked about?}}                                     & CoCoNot \\
\textbf{Type 2}: Undetermined lexicons or references     & {\textit{When was the last time the Giants went to the playoffs?}\\\textcolor{red}{"The Giants": the football team or the baseball team?}} & CLAMBER \\
\textbf{Type 3}: Undetermined perspective or granularity & {\textit{When was the First World War broke out?}\\\textcolor{red}{Is "broke out'' in the political sense or in the military sense?}}      & AmbigNQ \\
\textbf{Type 4}: Undetermined standard or preference     & {\textit{Recommend the best smartwatches available in 2023.}\\\textcolor{red}{What's the specific standard of being "best'"?}}             & IN3     
\end{tblr}

\caption{The taxonomic categories of underspecified queries in QA, with examples from multiple annotated datasets related to semantic underspecification.}
\label{tab: taxonomy}
\end{table}

\paragraph{Models} We experiment with \texttt{Qwen3} with four different parameter sizes (4B, 8B, 14B, and 32B) in the ``thinking mode'' \citep{yang2025qwen3}; \texttt{DeepSeek R1} \citep{DSR1report} distilled models with four parameter sizes (1.5B, 7B, 14B, and 32B); \texttt{Llama-3.2-3B-Instruct} \citep{llama3.2} and \texttt{Llama-3.3-70B-Instruct} \citep{meta2025llama3}.

\paragraph{Experimental setup} We prompt the LLMs via instructions to act as expert analysts and classify input questions as either FS or UND (UND covers various types of underspecification). The prompt includes a task description, the target question, and an explicit requirement for the output format.~\footnote{Refer to Appendix \ref{app: prompt} for the specific prompt used.} We analyze task accuracies and macro F1 values of all models on both UNDER and UNDER-gold to select the best-performing model.


\paragraph{Results} While several models perform comparably on the UND/FS classification task (see Table~\ref{tab: Exp 1 NL overview} in the Appendix \ref{app: results}), \texttt{Qwen3-4B} achieves the highest overall accuracy (0.71) and macro F1 (0.70) on the UNDER dataset. Its performance of 0.77 accuracy and 0.76 macro F1 on UNDER-gold also makes it the best among all tested models. Across both datasets, \texttt{Qwen3-4B} yields the highest average accuracy (0.74) and average macro F1 (0.73), outperforming the second best model \texttt{Qwen3-32B} (average accuracy 0.71, average macro F1 0.71) by a notable margin. Based on these results, we selected this model as the UND classifier for all subsequent experiments. These results indicate that this classifier satisfactorily distinguishes UND from FS questions, especially when labels align closely with expert annotations (i.e., UNDER-gold).



\begin{table*}[t] 
\centering
\small 
\resizebox{\textwidth}{!}{
\begin{tabular}{l|c|c|c|c}
\toprule
Models/Datasets & NQ UND $\rightarrow$ NQ Rewr & HotpotQA UND $\rightarrow$ HotpotQA Rewr & TriviaQA UND $\rightarrow$ TriviaQA Rewr & FRAMES UND $\rightarrow$ FRAMES Rewr \\
\midrule
GPT-4o & 37.0\% $\rightarrow$ 57.3\% (+20.3\%) & 34.6\% $\rightarrow$ 51.8\% (+17.2\%) & 75.8\% $\rightarrow$ 83.6\% (+7.8\%) & 24.4\% $\rightarrow$ 41.6\% (+17.2\%) \\
Gemini-2.5-Flash & 38.8\% $\rightarrow$ 50.0\% (+11.2\%) & 41.2\% $\rightarrow$ 50.6\% (+9.4\%) & 76.0\% $\rightarrow$ 74.4\% (-1.6\%) & 37.1\% $\rightarrow$ 46.5\% (+9.4\%) \\
\bottomrule
\end{tabular}
}
\caption{Comparison of F1 score for \texttt{GPT-4o} and \texttt{Gemini-2.5-Flash} across datasets. UND $\rightarrow$ Rewritten (+$\Delta$).}
\label{tbl:comparison}
\end{table*}

\subsection{Step 2: Assessing LLM QA performance}

\paragraph{Data} 
We experiment with four widely used QA datasets without underspecification-related annotations: Natural Questions (NQ) \citep{Kwiatkowski2019}, HotpotQA \citep{yang-etal-2018-hotpotqa}, TriviaQA \citep{joshi-etal-2017-triviaqa}, FRAMES \citep{krishna-etal-2025-fact}. In this experiment, we consider only the questions themselves, without providing any context passage to the model (e.g., the context paragraphs in HotpotQA). For FRAMES, we include all 824 $\langle$question, annotated answer$\rangle$ pairs. From each of the remaining datasets, we uniformly sample 1,000 instances, yielding a total of 3,824 data points. We refer to this aggregated collection as \textbf{QA-ensemble}.


\paragraph{Models}
We experiment with two proprietary LLMs, \texttt{GPT-4o} (\texttt{gpt-4o-2024-11-20}) \citep{openai2024gpt4o} and \texttt{Gemini-2.5-Flash} \citep{comanici2025gemini25pushingfrontier}, which we use as our \textit{QA models}.


\paragraph{Experimental setup}
We first apply the \textit{UND classifier} to QA-ensemble to automatically label questions as fully specified (FS) or underspecified (UND). The \textit{QA models} are then prompted to answer all questions, and performance is evaluated separately on the UND and FS subsets.\footnote{See Appendix~\ref{app: prompt} for the full QA prompt.} We assess performance using both a standard token-level F1 score and an LLM-as-judge metric, namely the Nvidia Answer Accuracy (AA) within the RAGAS framework~\citep{ragas2024,RAGAS-AA}.~\footnote{Nvidia AA measures alignment between a generated answer and a reference by averaging two independent LLM judgments on a discrete scale (0, 2, 4), normalized to $[0,1]$.} For each model, we conduct independent $t$-tests to determine whether performance on UND questions is significantly lower than on FS questions.

\paragraph{Results.}
\begin{figure}[t]
    \centering
    \includegraphics[width=\linewidth]{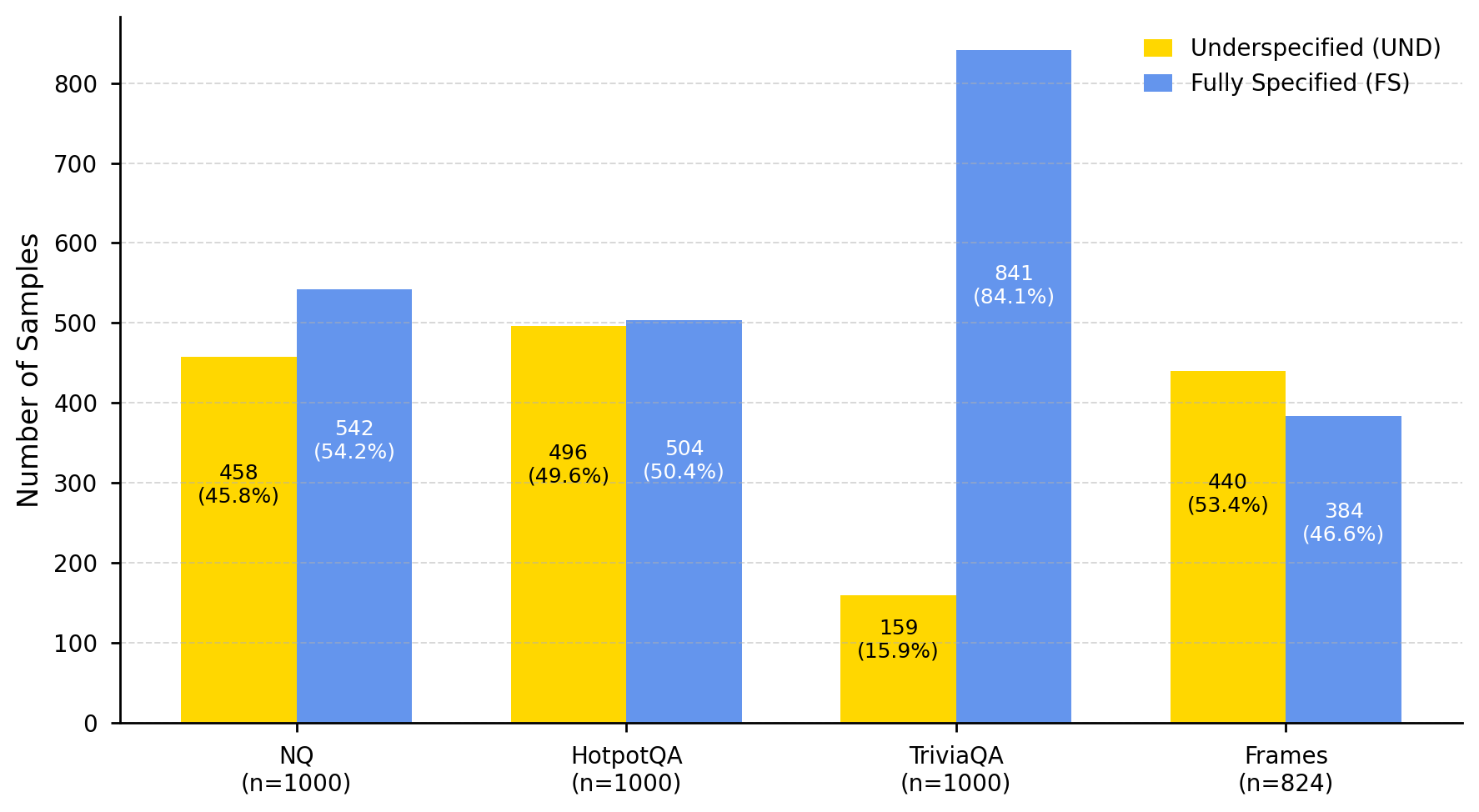}
    \caption{Proportion of FS/UND questions in each of the four source datasets included in QA-ensemble.}
    \label{fig: general QA classification}
\end{figure}

Fig.~\ref{fig: general QA classification} presents the FS/UND classification results of samples from the four source QA datasets provided by \textit{UND classifier}. The proportion of UND queries varies between datasets, with the lowest observed in TriviaQA (15.9\%) and the highest observed in FRAMES (53.4\%).
These results indicate a widespread presence of underspecified queries in all source QA datasets. By running \textit{QA models} on these partitions, we report a statistically significantly lower performance on UND compared to FS questions for all models and all datasets in QA-ensemble (see Appendix \ref{app: exp_2} for detailed results).




\subsection{Step 3: Rewriting UND questions}


\paragraph{Data and models}
We use all UND questions from QA-ensemble. We experiment with \texttt{GPT-4o} and \texttt{Gemini-2.5-Flash} as \textit{Rewriter models}.


\paragraph{Experimental setup}
We prompt the \textit{Rewriter models} to rewrite UND questions. The prompt includes golden answers to these UND queries annotated in their original QA datasets and the classification reasoning provided by the \textit{UND classifier} in Step 2 to supervise the rewriting task.\footnote{Please refer to Appendix \ref{app: prompt} for the rewriting prompt.} Then, we run the \textit{QA models} on these rewritten UND questions and report the percentage of queries that are classified as FS after the rewriting.

\paragraph{Results} The majority of rewritten questions are classified as fully specified, with proportions ranging from 64\% to over 86\%. This indicates that the rewriting step effectively removes underspecification, converting UND questions into FS ones. In the Appendix \ref{app: results}, Table~\ref{tab:Rewriting-Reclassification} reports the \textit{UND classifier}’s predictions on questions rewritten by our two \textit{Rewriter models}, \texttt{GPT-4o} and \texttt{Gemini-2.5-Flash}.  


\begin{table*}[!ht]
\captionsetup{width=\textwidth}
\centering
\tiny
\begin{tblr}{
  width = \linewidth,
  colspec = {Q[180]Q[60]Q[62]Q[170]Q[58]Q[60]Q[403]},
  cells = {c},
  cell{1}{1-7} = {Grandis},
  row{1} = {font=\tiny\bfseries},
  hlines,
  vlines,
  rowsep = 1pt,
  rows = {abovesep=2pt, belowsep=2pt},
}
Question & Source & QA Model & Golden Answers & {Model\\Answer} & {Qwen3-\\4B\\Pred} & Components leading to UND (Qwen3-4B) \\
{Who is the richest club in the championship?}
  & NQ & GPT-4o
  & {[`Manchester City']}
  & {[`Leicester City']}
  & UND
  & Underspecified standard: `richest'; undetermined reference:  `Championship' \\
{What was the code name of the landing barge primarily used to provide hot meals to the landing crew?}
  & HotpotQA & GPT-4o
  & {[`Operation Neptune']}
  & {[`HMS Men-estheus']}
  & UND
  & Potential references: a historically documented vessel, a fictional construct, or a specific operation? \\
{Who wrote the piece of music recognised in much of the Western world as ``The Wedding March'' (or as ``Here Comes The Bride'')?}
  & TriviaQA & Gemini-2.5-Flash
  & {[`Felix Mendelssohn', `mendelsson bartholdy' \ldots]}
  & {[`Richard Wagner']}
  & UND
  & Potential references: 1) `The Wedding March' from Wagner's \textit{Lohengrin}, 2) `Here Comes the Bride' from the 1935 film \textit{The Wizard of Oz}, 3) `Wedding March' from Mendelssohn's \textit{A Midsummer Night's Dream}. \\
{Who developed the first effective vaccine against the disease that killed the father of a famous Hungarian composer born in 1811?}
  & FRAMES & Gemini-2.5-Flash
  & {[Almroth Edward Wright]}
  & {[`Edward Jenner']}
  & UND
  & The specific disease causing the death remains ambiguous. Historical accounts suggest Franz Liszt Sr.\ died of typhoid fever, but this is contested. \\
\end{tblr}
\caption{Examples of UND queries with incorrect answers by \textit{QA models}}
\label{tab: unsatisfactory-UND-main}
\end{table*}

\subsection{Step 4: Reassessing QA performance}

\paragraph{Experimental Setup}
We reassess \textit{QA models} on the rewritten queries from Step 3 using the same QA prompt as in Step 2. To avoid bias from using the same LLM for both rewriting and answering, we adopt a cross-assignment setup: queries rewritten by \texttt{GPT-4o} are answered by \texttt{Gemini-2.5-Flash}, and \textit{vice versa}. We then compare QA performance on rewritten versus original UND queries using independent $t$-tests. 


\paragraph{Results}
As shown in Table~\ref{tbl:comparison}, \texttt{GPT-4o} and \texttt{Gemini-2.5-Flash} achieve broadly similar performance across all datasets. Rewriting queries to make them fully specified consistently improves results with the only exception of TriviaQA. This is expected since the dataset is explicitly designed to contain quiz questions that already provide sufficient information for answering. For example, \texttt{GPT-4o} on HotpotQA rises from 34.6\% to 51.8\% (+17.2\%), and \texttt{Gemini-2.5-Flash} on NQ rises from 38.8\% to 50.0\% (+11.2\%). These improvements indicate that when questions are well-formed and include all critical information, models can answer correctly, suggesting that the main source of errors in the UND setting is incomplete or unclear question formulation rather than limitations of the QA systems themselves. More detailed results with both F1 and Nvidia AA metrics are reported in in Appendix \ref{app: results}, Table~\ref{tab: Original_Rewritten QA Performance (full)}. Additionally, to rule out the potential effect of knowledge leakage where rewritten questions may contain vocabulary directly from golden answers, allowing models to exploit surface-level cues, we conduct a quantitative analysis to verify the lexical overlap between questions (original and rewritten) and their golden answers is minor. To do so, we use Jaccard similarity and n-gram overlap (unigram and bigram F1) as metrics and report the results in the Appendix \ref{app: results}, Table \ref{tab: Lexical overlap analysis}. This analysis shows that, while rewritten questions often exhibit statistically significantly higher lexical overlap with golden answers than original ones, absolute overlap values remain low ($<0.1$) across all metrics and configurations. This suggests that lexical similarities between rewritten questions and golden answers are marginal and unlikely to account for the observed performance gains.




\paragraph{Analysis}

We conduct a manual qualitative analysis of UND queries for which \textit{QA models} produce incorrect answers according to both F1 and NVIDIA AA metrics,\footnote{We consider samples with values $<$0.5 on both metrics.} with some examples shown in Table \ref{tab: unsatisfactory-UND-main}. We qualitatively observe that, in these cases, \textit{QA models} often generate answers that conflict with the gold annotation in the dataset, while this mismatch is resolved after rewriting. Additional examples are reported in Appendix~\ref{app: sheets}, Table \ref{tab: unsatisfactory UND} and Table \ref{tab: rewritten and corrected UND}.

\section{Conclusion}

Our study highlights the widespread presence of underspecified questions in current QA benchmarks, which can affect the reliability of QA model evaluation. We introduce an LLM-based classifier to detect such queries and a rewriter to transform them into fully specified forms. Across multiple widely used QA datasets, LLMs consistently struggle with underspecified questions when no supplementary contextual information (e.g., contextual paragraph) is provided, but performance improves significantly once underspecification/ambiguities are resolved, showing that many apparent QA failures reflect question clarity rather than model limitations. 

These findings highlight underspecification as a universally critical factor in QA evaluation, and we argue that future work should pay extra attention to QA underspecification by identifying, annotating, and possibly rewriting these questions. This will help shed light on the real capabilities of LLMs as effective and reliable QA models.

\section*{Limitations}
We acknowledge several limitations in this work. First, we adopted an off-the-shelf LLM as the UND-classifier. While its performance is reasonably strong, it is not optimal, and there remains room to improve classification accuracy and macro-F1 through prompt-based optimization or knowledge enhancement (e.g., RAG-based methods).

Second, implementation of our pipeline could be further improved by incorporating recent advances in LLM engineering, such as automatic prompt optimization and agentic frameworks.

Third, due to limited time and human resources, certain human expert verifications were not thoroughly conducted. For example, in Step 1, more manual inspection of the automatic mapping from original annotations to UND/FS labels could help mitigate cross-dataset inconsistencies in underspecification standards; in Step 3, comprehensive human review of rewritten questions could improve rewriting quality, recover failed cases, and prevent successfully rewritten questions from becoming overly trivial for downstream QA models.

Despite these limitations, we believe that our work provides a systematic and reproducible pipeline for detecting and rewriting underspecified QA queries, and that the main conclusions of our study remain robust.

\section*{Acknowledgments}
We would like to thank Jelke Bloem, Martha Lewis, and Malvin Gattinger for providing feedback on YH's master's thesis, which laid the foundations of this work. We are also grateful to the members of the \textit{Dialogue Modelling Group (DMG)} and the \textit{Multimodality, Language, and Interpretability (Mulini) lab} at the University of Amsterdam for their insightful feedback on the manuscript. We also acknowledge the use of AI assistants (Claude and ChatGPT) for code debugging in experiments.
\bibliography{custom}

\appendix
\section{A Working Taxonomy of Underspecified Queries in Question Answering (QA)}\label{app:taxonomy}

\paragraph{Type 1: Missing necessary components.} An underspecified query under this category contains at least one expression that is missing a commonly expected component conceptually tied to it (e.g., an implicit expected argument of a predicative element). Consequently, the semantic content of the expression at issue is linguistically incomplete and undertermined, with several different interpretations possible. This category is closely related to \textit{Missing Constituents} or \textit{Conceptual Truncation} discussed in the framework of Linguistic (Semantic) Underdeterminacy \citep{Carston2002,Belleri2014-BELSUA}.

\paragraph{Type 2: Undetermined lexicons or references.}  An underspecified query under this category contains at least one expression with lexical or referential ambiguity. Multiple same-level concepts or entities can be mapped to this expression at issue to serve as potential lexical entries or referents. It is impossible to fully determine which one is intended by the user based on the provided content. This type is closely related to \textit{lexical ambiguity} and \textit{referential ambiguity }\citep{Kennedy2011, sep-ambiguity,vanBerkum2008}, the concept of \textit{indexical reference} in the Linguistic Underdeterminacy framework \citep{Carston2002}, and the \textit{Syntactically but not semantically homogeneous ambiguities} discussed as a type of semantic underspecification in \citet{Egg2010}.

\paragraph{Type 3: Undetermined perspective or granularity.} An underspecified query under this category contains at least one expression where the general meaning is in place, but its specific interpretation can still vary based on different heterogeneous perspectives or granularity levels adopted. Multiple interpretations of different natures or levels are plausible for such an expression, and it's impossible to fully determine which one is intended based on the provided content.  As a result, the undeterminacy of this expression leads to multiple possible interpretations of the query, rendering it underspecified. Aligning with the view of \citet{Belleri2014-BELSUA}, we hold that it is not a type of lexical ambiguity as \citet{Kennedy2011, Kennedy2010-KENCCA-6} claims and it is also not a type of vagueness, as plausible perspectives or granularity levels are more definite and objectively acknowledged, instead of being completely a matter of contextual/subjective standard. We also claim that this category is diverging from prototypical cases of "missing constituents", as the linguistic intuition in cases under this category is more about inner interpretation of an expression at issue, instead of lacking external elements for the complete semantic saturation.

\paragraph{Type 4: Undetermined standard or preference.} An underspecified query under this category contains at least one expression where the general meaning is in place, but its specific interpretation is vague due to unspecified contextual standards or subjective criteria. A wide range of fine-grained interpretations is possible based on contextual or subjective needs, and it’s impossible to fully determine which one is intended by the user based on the provided content. As a result, the undeterminacy of this expression leads to many, or even an infinite number of, possible interpretations of the query. This category is closely related to the prototypical \textit{vagueness} \citet{Kennedy2011} and \textit{semantic imprecision} proposed by \citet{Bunt2007}. In the Linguistic (Semantic) Underdeterminacy framework, phenomena discussed under \textit{adjustments (overspecifying/underspecifying) of linguistically encoded concepts} \citep{Carston2002} and \textit{gradable expressions depending on standards or comparison classes} \citep{Belleri2014-BELSUA} can also be attributed to this category.

\section{Prompts}
\label{app: prompt}

\begin{tcolorbox}[breakable,mypromptbox]
\footnotesize{<SYSTEM\_PROMPT> You are an expert analyst. Your task is to analyze and determine whether an input user query is "fully specified" or "underspecified". </SYSTEM\_PROMPT>\\

Analyze the following input user query:\\

\{"query": "When did the nuclear accident happen?"\}\\
 
Please provide your analysis in the following JSON format:\\
 
\{"query": "When did the nuclear accident happen?", \\"reasoning": "[YOUR\_DETAILED\_REASONING]",\\ "judgment": "[fully specified/underspecified]"\}}
\end{tcolorbox}

\begin{tcolorbox}[breakable,mypromptbox16]
\footnotesize{<SYSTEM\_PROMPT> Answer the question with a concise response. Return answers as a list of strings. If there's only one answer, return a single-item list. Each answer should be brief and direct. </SYSTEM\_PROMPT>\\
{"role": "system", "content": [SYSTEM\_PROMPT]},\\
{"role": "user", "content": [QUESTION]}}
\end{tcolorbox}

\begin{tcolorbox}[breakable,mypromptbox2]
\footnotesize{<SYSTEM\_PROMPT>You are a professional question optimization expert. Please modify the underspecified question to a fully specified version based on the provided clues.\\
        Requirements:\\
        1. Keep the core intent of the question unchanged\\
        2. Add necessary contextual information\\
        3. Eliminate underspecified elements and make the question clear\\
        4. Ensure the modified question can be directly answered with the provided short answer without dispute\\
        Please only return the modified question, do not include any other explanations. </SYSTEM\_PROMPT>
        \\
The original question: {[QUESTION]}\\
Short answer: {[GOLD\_ANSWER]}\\
Reasoning: {[MODEL\_CLASSIFIER\_REASONING]}\\

Please analyze the underspecified elements in the original question, then modify the question to a fully specified version based on the short answer and reasoning.}
\end{tcolorbox}

\section{Results} \label{app: results}
Please refer to Tables \ref{tab: Exp 1 NL overview} - \ref{tab: Lexical overlap analysis}.

\begin{table*}[t]
\centering
\scriptsize
\begin{tblr}{
  width = \linewidth,
  colspec = {Q[100]Q[105]Q[100]Q[40]Q[105]Q[100]Q[40]Q[105]Q[100]Q[40]Q[105]Q[100]Q[40]Q[105]Q[100]Q[40]},
  cells = {c},
  cell{1}{1} = {r=2}{},
  cell{1}{2} = {c=3}{Grandis},
  cell{1}{5} = {c=3}{Grandis},
  cell{1}{8} = {c=3}{Grandis},
  cell{1}{11} = {c=3}{Grandis},
  cell{1}{14} = {c=3}{Grandis},
  cell{3}{2} = {font=\bfseries},
  cell{3}{3} = {font=\bfseries},
  cell{3}{4} = {font=\bfseries},
  cell{4}{2} = {font=\bfseries},
  cell{4}{3} = {font=\bfseries},
  cell{4}{4} = {font=\bfseries},
  cell{4}{8} = {font=\bfseries},
  cell{4}{11} = {font=\bfseries},
  cell{5}{2} = {c=3}{Grandis},
  cell{5}{5} = {c=3}{Grandis},
  cell{5}{8} = {c=3}{Grandis},
  cell{5}{11} = {c=3}{Grandis},
  cell{5}{14} = {c=3}{Grandis},
  cell{7}{14} = {font=\bfseries},
  cell{8}{14} = {font=\bfseries},
  hlines,
  vlines,
}
         & Qwen3-4B   &            &      & Qwen3-8B &            &      & Qwen3-14B &            &      & Qwen3-32B &            &      & Llama-3.2-3B  &            &      \\
         & UNDER      & UNDER-gold & Avg  & UNDER    & UNDER-gold & Avg  & UNDER     & UNDER-gold & Avg  & UNDER     & UNDER-gold & Avg  & UNDER         & UNDER-gold & Avg  \\
accuracy & 0.71       & 0.77       & 0.74 & 0.68     & 0.71       & 0.70 & 0.70      & 0.67       & 0.69 & 0.70      & 0.72       & 0.71 & 0.51          & 0.71       & 0.61 \\
macro F1 & 0.70       & 0.76       & 0.73 & 0.67     & 0.69       & 0.68 & 0.70      & 0.66       & 0.68 & 0.70      & 0.71       & 0.71 & 0.38          & 0.56       & 0.47 \\
         & DS-R1-1.5B &            &      & DS-R1-7B &            &      & DS-R1-14B &            &      & DS-R1-32B &            &      & Llama-3.3-70B &            &      \\
         & UNDER      & UNDER-gold & Avg  & UNDER    & UNDER-gold & Avg  & UNDER     & UNDER-gold & Avg  & UNDER     & UNDER-gold & Avg  & UNDER         & UNDER-gold & Avg  \\
accuracy & 0.55       & 0.66       & 0.61 & 0.62     & 0.74       & 0.68 & 0.68      & 0.68       & 0.68 & 0.67      & 0.73       & 0.70 & 0.71          & 0.68       & 0.70 \\
macro F1 & 0.49       & 0.56       & 0.53 & 0.62     & 0.73       & 0.68 & 0.68      & 0.66       & 0.67 & 0.66      & 0.70       & 0.68 & 0.70          & 0.68       & 0.69 
\end{tblr}
\caption{An overview of the performance on \textbf{UNDER} and \textbf{UNDER-gold} across the selected LLMs.}
\label{tab: Exp 1 NL overview}
\end{table*}

\begin{table*}[t]
\captionsetup{width=\textwidth}
\centering
\scriptsize
\begin{tblr}{
  width = 0.9\linewidth,
  colspec = {X[0.35] X[0.67] X[0.70] X[0.67] X[0.70] X[0.67] X[0.70] X[0.67] X[0.70]},
  cells = {c},
  cell{1}{1} = {r=2}{},
  cell{1}{2} = {c=2}{Grandis},
  cell{1}{4} = {c=2}{Grandis},
  cell{1}{6} = {c=2}{Grandis},
  cell{1}{8} = {c=2}{Grandis},
  vlines,
  hline{1,3,4} = {-}{},   
  hline{2} = {2-9}{},
}
 & NQ-Rewr &  & HotpotQA-Rewr &  & TriviaQA-Rewr &  & FRAMES-Rewr &  \\
 & GPT-4o & Gemini-2.5-Flash & GPT-4o & Gemini-2.5-Flash & GPT-4o & Gemini-2.5-Flash & GPT-4o & Gemini-2.5-Flash \\
FS
 & 85.4\% & 84.5\%
 & 72.2\% & 83.6\%
 & 86.1\% & 83.6\%
 & 86.2\% & 83.6\%
\end{tblr}
\caption{For each dataset in QA-ensemble: Percentage of queries rewritten by the \textit{Rewriter model} (either \texttt{GPT-4o} or \texttt{Gemini-2.5-Flash}) which are classified as FS by the \textit{UND classifier}. This proportion is high: 64--86\%.}
\label{tab:Rewriting-Reclassification}
\end{table*}

\begin{table*}[t]
\captionsetup{width=\textwidth}
\centering
\scriptsize
\begin{tblr}{
  width = \linewidth,
  colspec = {Q[200]Q[87]Q[70]Q[87]Q[70]Q[87]Q[70]Q[87]Q[70]},
  cell{1}{1} = {r=2}{},
  cell{1}{2} = {c=2}{0.174\linewidth,Grandis},
  cell{1}{4} = {c=2}{0.174\linewidth,Grandis},
  cell{1}{6} = {c=2}{0.174\linewidth,Grandis},
  cell{1}{8} = {c=2}{0.174\linewidth,Grandis},
  vlines,
  hline{1,3-7} = {-}{},
  hline{2} = {2-9}{},
}
                    & NQ     &        & HotpotQA &        & TriviaQA &        & FRAMES &        \\
                    & Orig UND    & Rewr     & Orig UND      & Rewr     & Orig UND      & Rewr     & Orig UND    & Rewr     \\
GPT-4o F1           & 37.0\% & 57.3\% & 34.6\%   & 51.8\% & 75.8\%   & 83.6\% & 24.4\% & 41.6\% \\
GPT-4o AA           & 58.0\% & 73.4\% & 42.2\%   & 61.3\% & 84.3\%   & 92.8\% & 27.2\% & 51.5\% \\
Gemini-2.5-Flash F1 & 38.8\% & 50.0\% & 41.2\%   & 50.6\% & 76.0\%   & 74.4\% & 37.1\% & 46.5\% \\
Gemini-2.5-Flash AA & 54.9\% & 68.1\% & 45.3\%   & 60.5\% & 84.3\%   & 87.6\% & 42.0\% & 54.6\% 
\end{tblr}
\caption{The performance of QA models on original UND queries and their rewritten counterparts across QA datasets. }
\label{tab: Original_Rewritten QA Performance (full)}
\end{table*}

\begin{table*}[t]
\captionsetup{width=\textwidth}
\centering
\scriptsize
\begin{tblr}{
  width = \linewidth,
  colspec = {Q[155]Q[60]Q[60]Q[60]Q[85]Q[60]Q[60]Q[60]Q[85]Q[60]Q[60]Q[80]Q[85]},
  cells = {c},
  row{3} = {Pippin},
  row{6} = {Pippin},
  row{9} = {Pippin},
  row{12} = {Pippin},
  cell{1}{2} = {c=4}{Grandis},
cell{1}{6} = {c=4}{Grandis},
cell{1}{10} = {c=4}{Grandis},
cell{3}{1} = {c=13}{},
cell{6}{1} = {c=13}{},
cell{9}{1} = {c=13}{},
cell{12}{1} = {c=13}{},
  hlines,
  vlines,
}
                     & Jaccard &        &        &              & Unigram F1 &        &        &              & Bigram F1 &        &         &              \\
datasets + QA models & Orig    & Rewr   & $\Delta$      & Sig & Orig       & Rewr   & $\Delta$      & Sig & Orig      & Rewr   & $\Delta$       & Sig \\
NQ                   &         &        &        &              &            &        &        &              &           &        &         &              \\
GPT-4o               & 0.0293  & 0.0477 & 0.0184 & p < 0.001      & 0.0488     & 0.077  & 0.0282 & p < 0.001      & 0.0049    & 0.0196 & 0.0148  & p < 0.001      \\
Gemini-2.5-Flash     & 0.0293  & 0.0711 & 0.0418 & p < 0.001      & 0.0488     & 0.1074 & 0.0586 & p < 0.001      & 0.0049    & 0.0512 & 0.0463  & p < 0.001      \\
HotpotQA             &         &        &        &              &            &        &        &              &           &        &         &              \\
GPT-4o               & 0.0368  & 0.0379 & 0.0011 & p=0.8828     & 0.0606     & 0.0611 & 0.0005 & p=0.7418     & 0.0204    & 0.0203 & -0.0001 & p=0.3560     \\
Gemini-2.5-Flash     & 0.0368  & 0.0648 & 0.028  & p < 0.001      & 0.0606     & 0.1034 & 0.0428 & p < 0.001      & 0.0204    & 0.0535 & 0.0331  & p < 0.001      \\
TriviaQA             &         &        &        &              &            &        &        &              &           &        &         &              \\
GPT-4o               & 0.0203  & 0.0319 & 0.0116 & p < 0.001      & 0.0254     & 0.0417 & 0.0164 & p < 0.001      & 0.0038    & 0.0076 & 0.0037  & p  0.01      \\
Gemini-2.5-Flash     & 0.0203  & 0.0471 & 0.0267 & p < 0.001      & 0.0254     & 0.0662 & 0.0409 & p < 0.001      & 0.0038    & 0.0187 & 0.0149  & p < 0.001      \\
FRAMES               &         &        &        &              &            &        &        &              &           &        &         &              \\
GPT-4o               & 0.0506  & 0.0656 & 0.0151 & p < 0.001      & 0.0725     & 0.0912 & 0.0187 & p < 0.001      & 0.0312    & 0.0475 & 0.0164  & p < 0.001      \\
Gemini-2.5-Flash     & 0.0506  & 0.07   & 0.0194 & p < 0.001      & 0.0725     & 0.1006 & 0.0281 & p < 0.001      & 0.0312    & 0.0503 & 0.0191  & p < 0.001      
\end{tblr}
\caption{The lexical overlap analysis between questions (original and rewritten) and their golden answers. "Orig" stands for lexical overlap between original questions and golden answers; "Rewr" stands for the lexical overlap between rewritten questions and golden answers; $\Delta$ represents the numeric gap between two measurements; "Sig" is the statistical significance obtained via the Wilcoxon signed-rank test. }
\label{tab: Lexical overlap analysis}
\end{table*}

\section{Figures}

\subsection{Step 2: UND-FS Performance Comparisons of Models-QA} \label{app: exp_2}
\begin{figure*}[h]
    \captionsetup{width=\textwidth}
    \centering
    \includegraphics[width=0.8\linewidth]{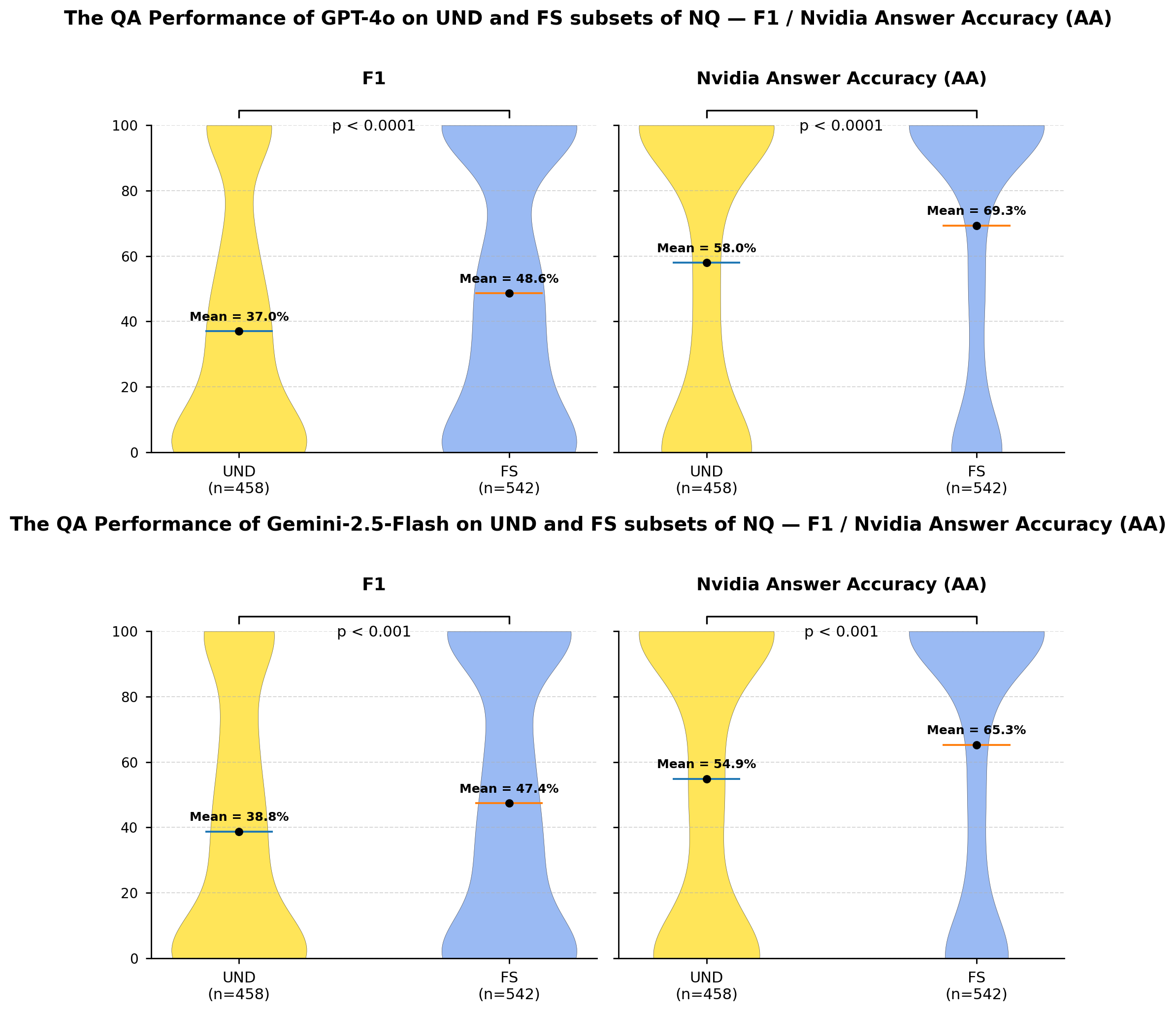}
    \caption{(Step 2) The FS/UND classification results of the \textbf{NQ} sample in \textbf{QA-ensemble}.}
    \label{fig: UND-FS-Comparisons-NQ-B}
\end{figure*}

\begin{figure*}[h]
    \captionsetup{width=\textwidth}
    \centering
    \includegraphics[width=0.8\linewidth]{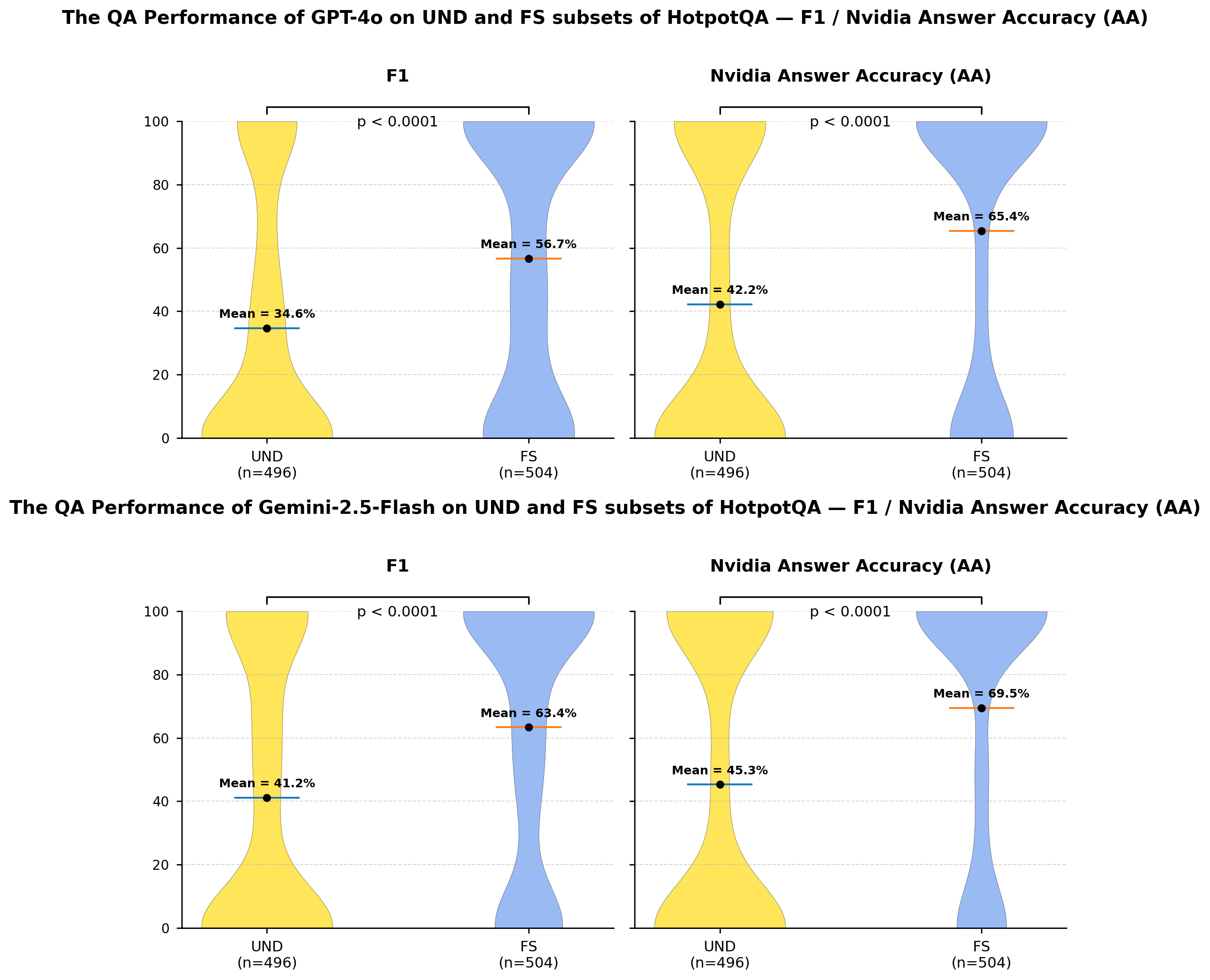}
    \caption{(Step 2) The FS/UND classification results of the \textbf{HotpotQA} sample in \textbf{QA-ensemble}.}
    \label{fig: UND-FS-Comparisons-HotpotQA-B}
\end{figure*}

\begin{figure*}[h]
    \captionsetup{width=\textwidth}
    \centering
    \includegraphics[width=0.8\linewidth]{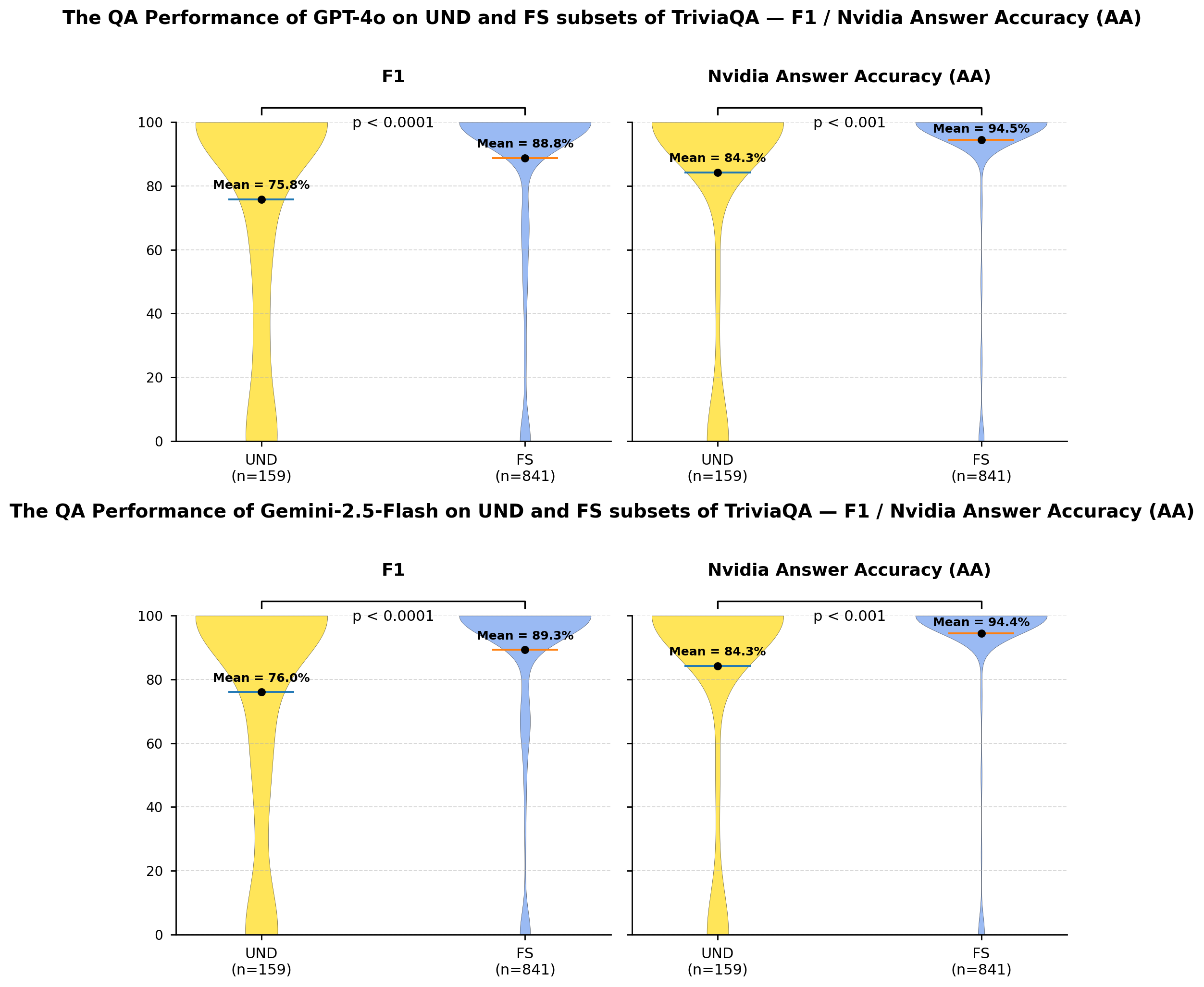}
    \caption{(Step 2) The FS/UND classification results of the \textbf{TriviaQA} sample in \textbf{QA-ensemble}.}
    \label{fig: UND-FS-Comparisons-TriviaQA-B}
\end{figure*}

\begin{figure*}[h]
    \captionsetup{width=\textwidth}
    \centering
    \includegraphics[width=0.8\linewidth]{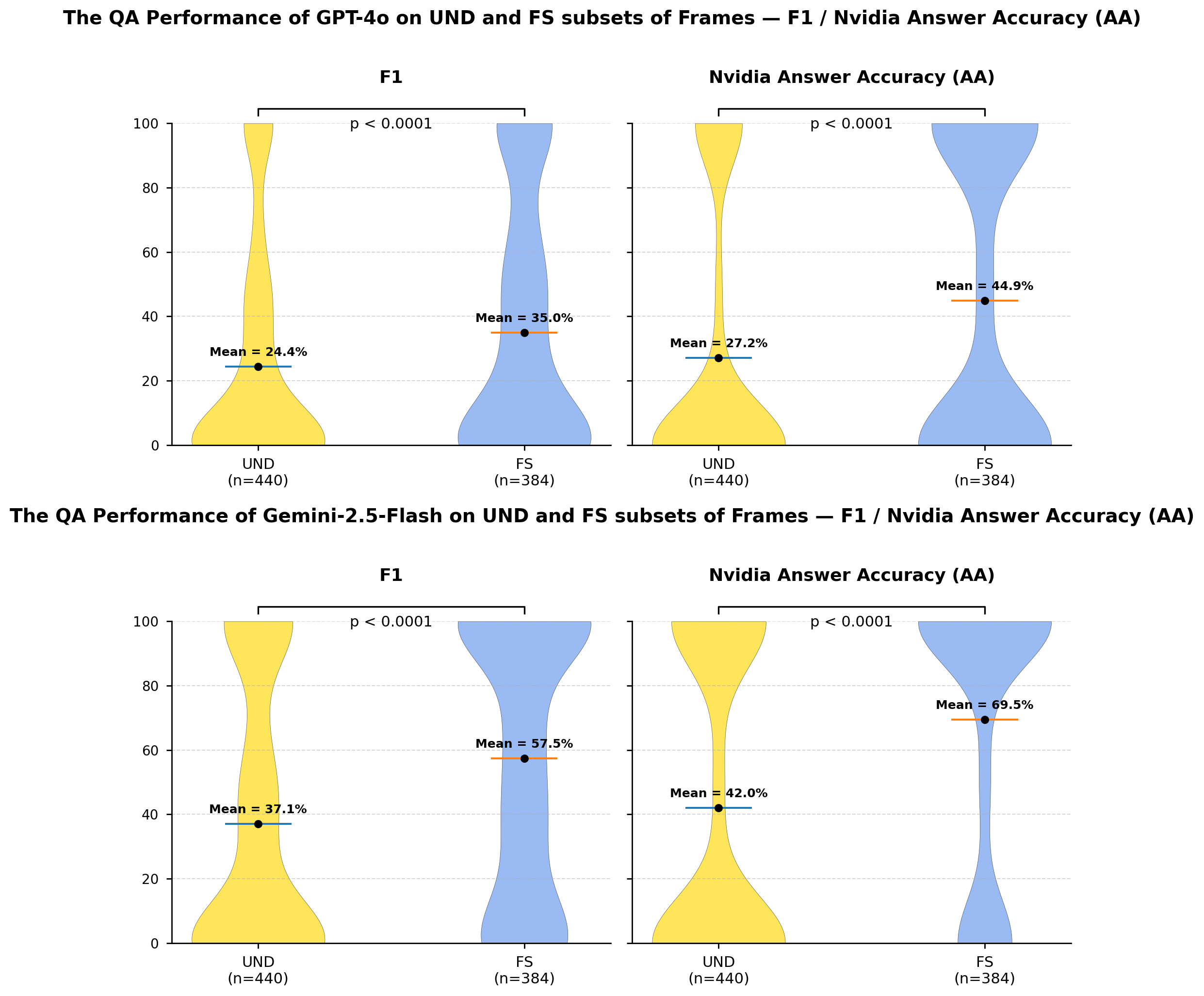}
    \caption{(Step 2) The FS/UND classification results of the \textbf{FRAMES} sample in \textbf{QA-ensemble}.}
    \label{fig: UND-FS-Comparisons-Frames-B}
\end{figure*}

Please refer to Fig.~\ref{fig: UND-FS-Comparisons-NQ-B} - \ref{fig: UND-FS-Comparisons-Frames-B}.

\subsection{Step 4: Original-Rewritten Performance Comparisons of Models QA} \label{app: exp_3}
\begin{figure*}[h]
    \captionsetup{width=\textwidth}
    \centering
    \includegraphics[width=0.7\linewidth]{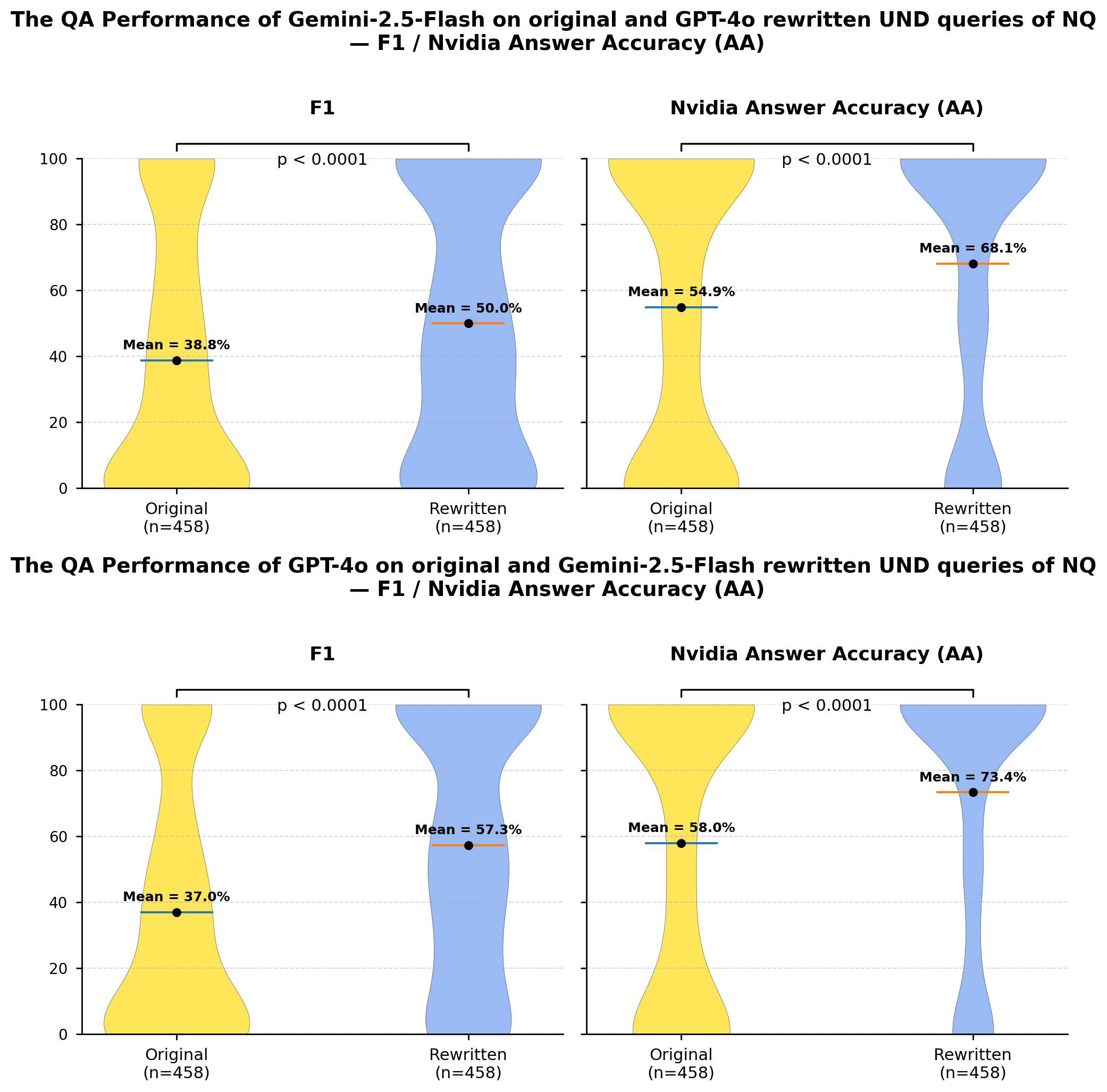}
    \caption{(Step 4) Comparing classification results of the original UND queries and their rewritten counterparts from the \textbf{NQ} sample in \textbf{QA-ensemble}.}
    \label{fig: NQ-rewrite}
\end{figure*}

\begin{figure*}[h]
    \captionsetup{width=\textwidth}
    \centering
    \includegraphics[width=0.7\linewidth]{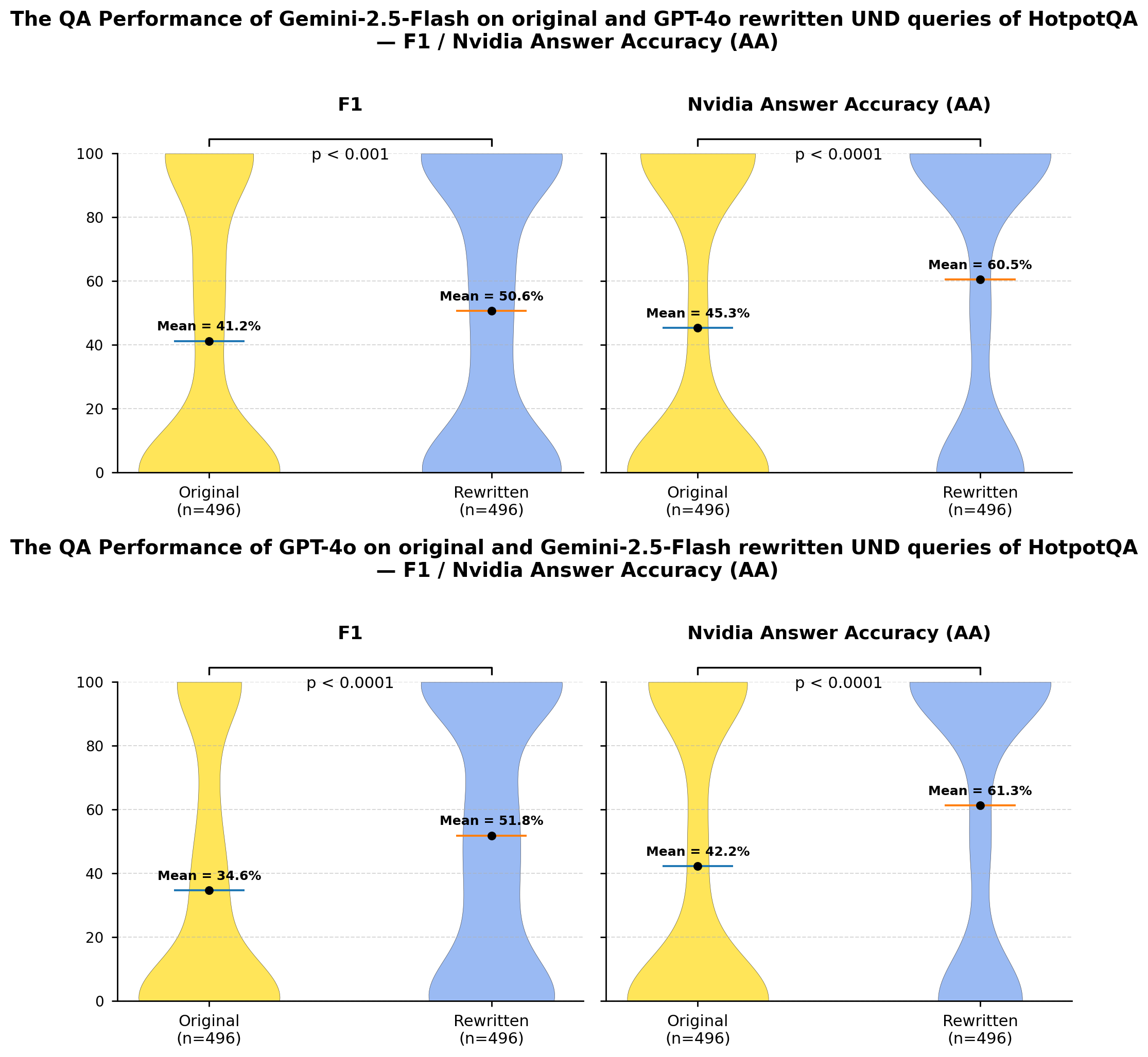}
    \caption{(Step 4) Comparing classification results of the original UND queries and their rewritten counterparts from the \textbf{HotpotQA} sample in \textbf{QA-ensemble}.}
    \label{fig: HotpotQA-rewrite}
\end{figure*}

\begin{figure*}[h]
    \captionsetup{width=\textwidth}
    \centering
    \includegraphics[width=0.7\linewidth]{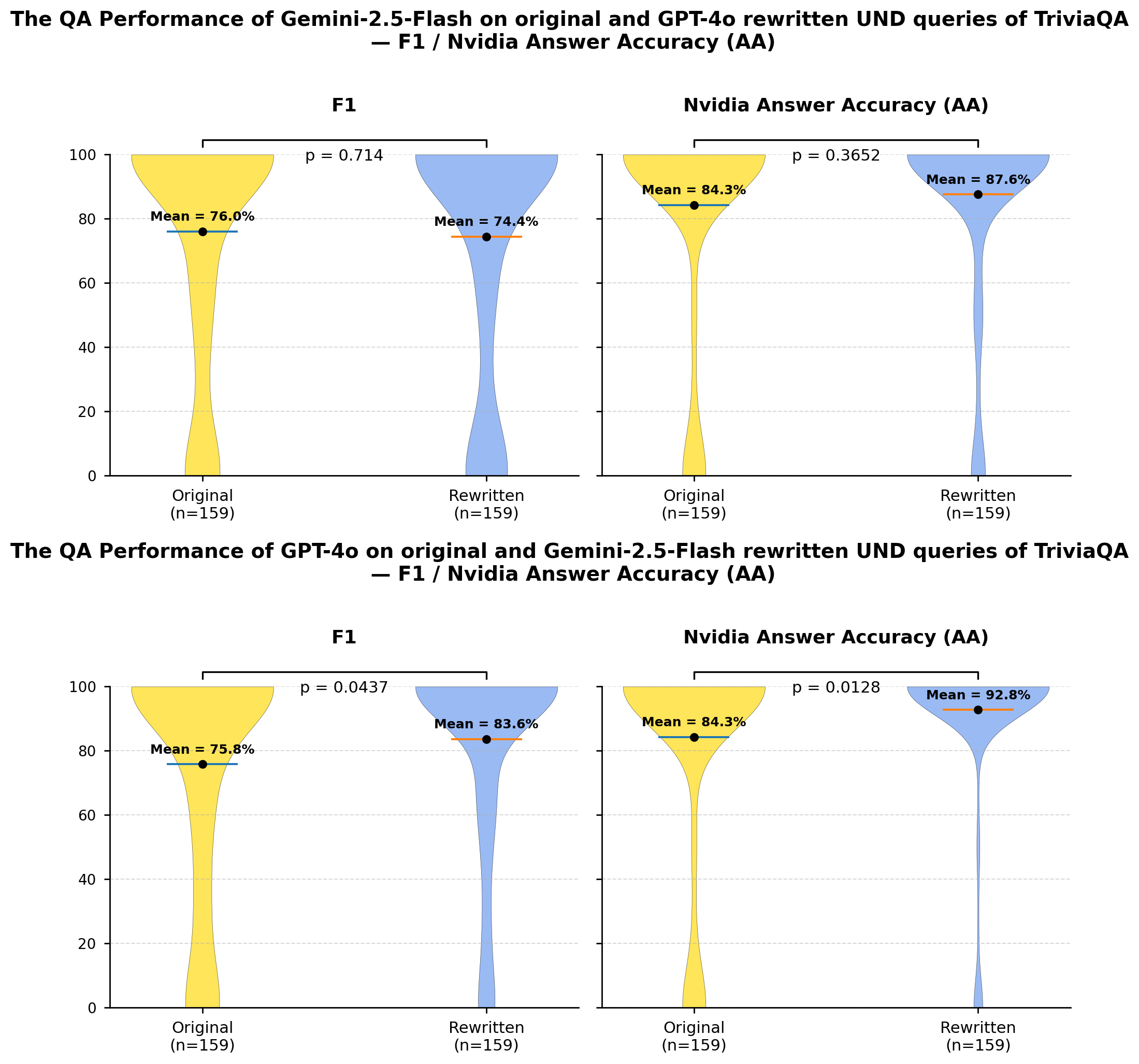}
    \caption{(Step 4) Comparing classification results of the original UND queries and their rewritten counterparts from the \textbf{TriviaQA} sample in \textbf{QA-ensemble}.}
    \label{fig: TriviaQA-rewrite}
\end{figure*}

\begin{figure*}[h]
    \captionsetup{width=\textwidth}
    \centering
    \includegraphics[width=0.7\linewidth]{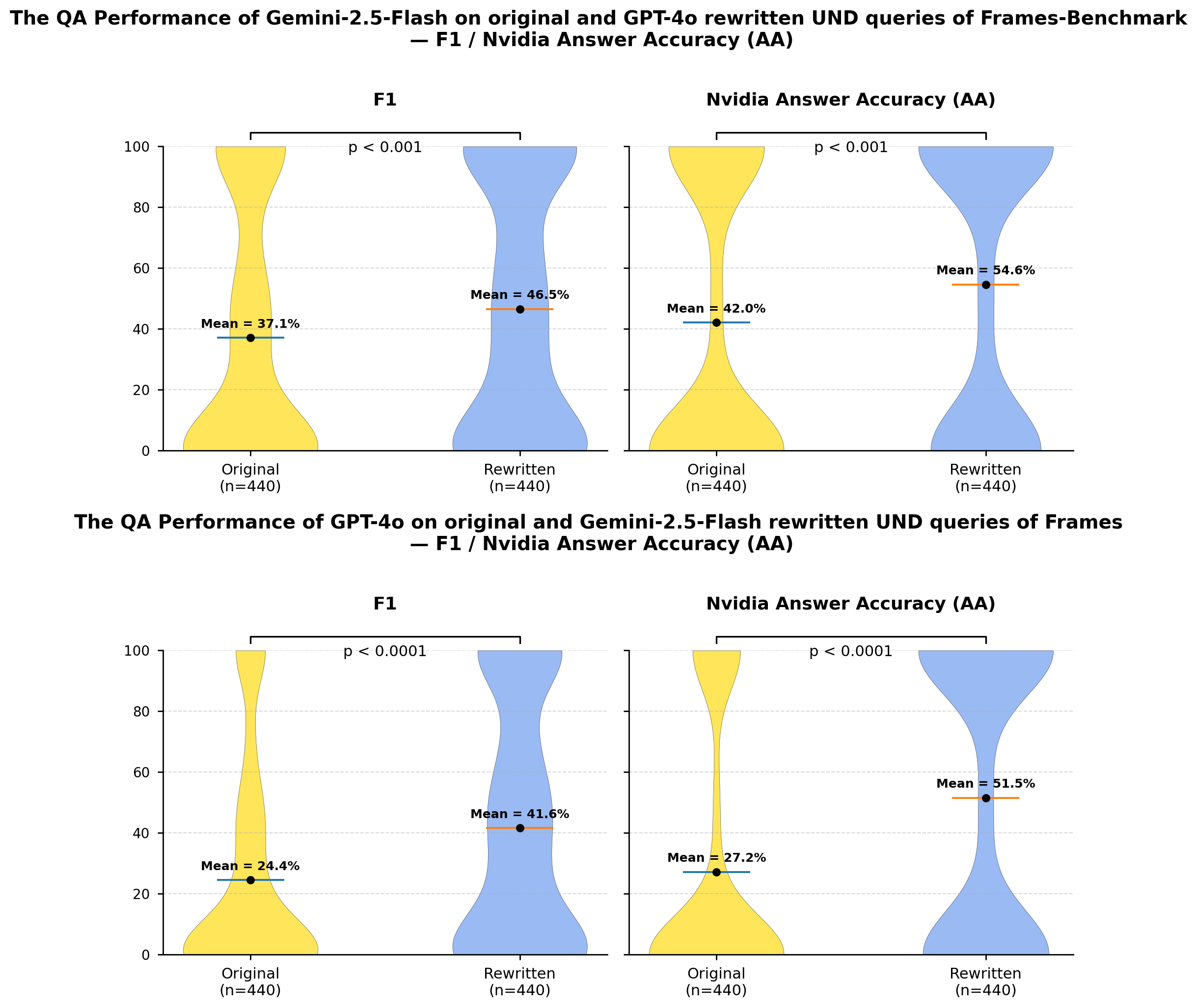}
    \caption{(Step 4) Comparing classification results of the original UND queries and their rewritten counterparts from the \textbf{FRAMES} sample in \textbf{QA-ensemble}.}
    \label{fig: Frames-rewrite}
\end{figure*}

Please refer to Fig.~\ref{fig: NQ-rewrite} - \ref{fig: Frames-rewrite}.

\section{More Selected Examples from Qualitative Analysis}\label{app: sheets}
\begin{itemize}
    \item Please refer to Table \ref{tab: unsatisfactory UND} for selected examples of UND queries with incorrect answers by \textit{QA models}.
    \item  Please refer to Table \ref{tab: rewritten and corrected UND} for selected examples of UND queries rewritten by \textit{Rewriter} models, which were initially answered incorrectly by \textit{QA models} but answered correctly in a subsequent round after rewriting.
\end{itemize}

\begin{table*}[t]
\captionsetup{width=\textwidth}
\centering
\scriptsize
\begin{tblr}{
  width = \linewidth,
  colspec = {Q[158]Q[80]Q[80]Q[140]Q[100]Q[42]Q[371]},
  cells = {c},
  row{1} = {Grandis},
  hlines,
  vlines,
}
Question                                                                                                                          & Source   & QA Model         & Golden Answers                                                                                                                    & Model Answer                                                                                                                                & Qwen3-4B Pred & Qwen3-4B reasoning                                                                                                                                                                                                                                                                                                                                                                                                                                         \\
Who is the richest club in the championship?                                                                                      & NQ       & GPT-4o           & {[}'Manchester City']                                                                                                           & {[}'Leicester City']                                                                                                                        & UND           & The query is ambiguous because 'richest' is not defined (e.g., revenue, profit, asset value). The term 'Championship' could refer to different leagues globally (e.g., English FA Cup, Spanish LaLiga, etc.)... Furthermore, the definition of 'richest' varies by metric (broadcast income vs. operating profit vs. shareholder equity)...                                                                                                                \\
Who owns the majority of the US debt?                                                                                             & NQ       & Gemini-2.5-Flash & {[}'Japan and China']                                                                                                           & {[}'* The U.S. government itself (through intragovernmental holdings)']                                                                     & UND           & ...The term 'majority' introduces ambiguity. Key considerations include: 1) Whether 'majority' refers to the largest single holder (e.g., Foreign Governments, Federal Reserve, or Domestic Institutions) or aggregate holdings; 2) Timeframe (current vs historical); 3) Debt instrument specificity (Treasury Bonds, Notes, Bills)...                                                                                                                    \\
What was the code name of the landing barge primarily used to provide hot meals to the landing crew?                              & HotpotQA & GPT-4o           & {[}'Operation Neptune']                                                                                                         & {[}'HMS Menestheus']                                                                                                                        & UND           & While 'landing barge' is a technical term for a type of watercraft used in amphibious operations, the phrase 'The landing barge' implies a singular, named entity. However, the query does not clarify whether this refers to a historically documented vessel, a fictional construct, or a specific operation...                                                                                                                                          \\
What colloquial umbrella do Rutgers and the many Ivy league schools fall under?                                                   & HotpotQA & Gemini-2.5-Flash & {[}'colonial colleges']                                                                                                         & {[}'* Elite schools\textbackslash{}n* Top-tier universities\textbackslash{}n* Ivy-level schools']                                           & UND           & The phrase 'colloquial umbrella' suggests seeking a shared informal label, but Rutgers and Ivy League schools belong to distinct categorizations (public vs. private, regional affiliations). The contradiction between Rutgers (non-Ivy) and Ivy League schools creates ambiguity about what unifying term the user seeks.                                                                                                                                \\
Which month of the year came second in the Roman calendar?                                                                        & TriviaQA & GPT-4o           & {[}'april month' 'april' 'mali traven' ...]                          & {[}'February']                                                                                                                              & UND           & ...The query asks for the second month in the 'Roman calendar,' but does not clarify which specific version of the Roman calendar being referenced. The ancient Roman calendar initially began with March as the first month, making February the second. Later reforms, such as the Julian calendar introduced by Julius Caesar, reorganized the calendar to start with January as the first month, making February the second again...                   \\
Who wrote the piece of music recognised in much of the Western world as ""The Wedding March"" (or as ""Here Comes The Bride"")?   & TriviaQA & Gemini-2.5-Flash & {[}'felix mendelssohn' 'mendelsson bartholdy'...] & {[}'* Richard Wagner']                                                                                                                      & UND           & The query asks about the author of a piece known as 'The Wedding March' or 'Here Comes The Bride.' While these terms are widely recognized, they refer to multiple compositions across different eras and genres. For example: 1) 'The Wedding March' from Wagner's *Lohengrin*, 2) 'Here Comes the Bride' from the 1935 film *The Wizard of Oz* (based on a 1740 French melody), and 3) 'Wedding March' from Mendelssohn's *A Midsummer Night's Dream*... \\
Who was the King of Siam during the 6th deadliest single-day terrorist attack in U.S. history?                                    & FRAMES   & GPT-4o           & {[}King Prajadhipok]                                                                                                            & {[}'There was no King of Siam during the 6th deadliest single-day terrorist attack in U.S. history, as Siam was renamed Thailand in 1939.'] & UND           & ... The reference to 'the 6th deadliest single-day terrorist attack in U.S. history' is ambiguous. First, the ranking of terrorist attacks by lethality is subjective and subject to change based on data sources (e.g., differing definitions of 'terrorist attack'). Second, even if the attack's date were determined, Siam's monarchy changed over time—Siam became Thailand in 1939—and the king would vary depending on the year of the attack.      \\
Who developed the first effective vaccine against the disease that killed the father of a famous Hungarian composer born in 1811? & FRAMES   & Gemini-2.5-Flash & {[}Almroth Edward Wright]                                                                                                       & {[}'Edward Jenner']                                                                                                                         & UND           & ..While the composer and his father’s identity are relatively clear (Franz Liszt and his father, Franz Liszt Sr.), the specific disease causing his death remains ambiguous. Historical accounts suggest Franz Liszt Sr. died of typhoid fever, but this is contested. Even if correct, the 'first effective vaccine' for typhoid predates modern times, complicating the answer. ..                                                                       
\end{tblr}
\caption{Additional selected examples of UND queries with incorrect answers by \textit{QA models}}
\label{tab: unsatisfactory UND}
\end{table*}

\begin{table*}[t]
\captionsetup{width=\textwidth}
\centering
\scriptsize
\begin{tblr}{
  width = \linewidth,
  colspec = {Q[92]Q[190]Q[100]Q[230]Q[150]Q[100]},
  cells = {c},
  row{1} = {Grandis},
  hlines,
  vlines,
}
Config                                          & Orig Q                                                                                                                            & Golden A~                                           & Rewr Q                                                                                                                                                                                                                                 & Orig A                                                                                                                                      & Rewr A                            \\
From NQ, GPT-4o Rewr, Gemini-2.5-Flash QA       & Who owns the majority of the US debt?                                                                                             & {[}'Japan and China']                               & Which foreign countries currently hold the largest portions of U.S. federal debt, specifically focusing on aggregate holdings, and who are the top two holders?                                                                        & {[}'* The U.S. government itself (through intragovernmental holdings)']                                                                     & {[}'Japan,~ China']               \\
From NQ, Gemini-2.5-Flash Rewr, GPT-4o QA       & Who is the richest club in the championship?                                                                                      & {[}'Manchester City']                               & Based on the most recent financial reports, which club in the English Premier League generated the highest revenue?                                                                                                                    & {[}'Leicester City']                                                                                                                        & {[}'Manchester City']             \\
From HotpotQA, GPT-4o Rewr, Gemini-2.5-Flash QA & What colloquial umbrella do Rutgers and the many Ivy league schools fall under?                                                   & {[}'colonial colleges']                             & What historical category do Rutgers University and the Ivy League schools collectively belong to, based on their founding during the colonial period?                                                                                  & {[}Elite schools / Top-tier universities / Ivy-level schools']                                                                              & {[}'Colonial colleges']           \\
From HotpotQA, Gemini-2.5-Flash Rewr, GPT-4o QA & What was the code name of The landing barge primarily used to provide hot meals to the landing crew?                              & {[}'Operation Neptune']                             & What was the code name for the naval assault phase of the Normandy landings, which encompassed the deployment of numerous landing barges and extensive logistical efforts, including the provision of hot meals to the landing forces? & {[}'HMS Menestheus']                                                                                                                        & {[}'Operation Neptune']           \\
From TriviaQA, GPT-4o Rewr, Gemini-2.5-Flash QA & Who wrote the piece of music recognised in much of the Western world as "The Wedding March""(or as "Here Comes The Bride")?       & {[}'felix mendelssohn'/ 'mendelsson bartholdy'/...] & Who composed the 'Wedding March' from *A Midsummer Night's Dream*, which is widely recognized in much of the Western world as 'The Wedding March'?                                                                                     & {[}'* Richard Wagner']                                                                                                                      & {[}'Felix Mendelssohn']           \\
From TriviaQA, Gemini-2.5-Flash Rewr, GPT-4o QA & Which month of the year came second in the Roman calendar?                                                                        & {[}''april month' / 'april' / ...]                  & In the early Roman calendar, when March was considered the first month of the year, which month came second?                                                                                                                           & {[}'February']                                                                                                                              & {[}'April']                       \\
From FRAMES, GPT-4o Rewr, Gemini-2.5-Flash QA   & Who developed the first effective vaccine against the disease that killed the father of a famous Hungarian composer born in 1811? & Almroth Edward Wright                               & Who developed the first effective vaccine against typhoid fever, the disease historically believed to have caused the death of Franz Liszt's father, Franz Liszt Sr.?                                                                  & {[}'Rudolf Weigl']                                                                                                                          & {[}'Almroth Wright']              \\
From FRAMES, Gemini-2.5-Flash Rewr, GPT-4o QA   & Who was the King of Siam during the 6th deadliest single-day terrorist attack in U.S. history?                                    & King Prajadhipok                                    & Who was the King of Siam during the Bath School bombing on May 18, 1927?                                                                                                                                                               & {[}'There was no King of Siam during the 6th deadliest single-day terrorist attack in U.S. history, as Siam was renamed Thailand in 1939.'] & {[}'King Prajadhipok (Rama VII)'] 
\end{tblr}
\caption{
Selected examples of UND queries rewritten by \textit{Rewriter} models, which were initially answered incorrectly by \textit{QA models} but answered correctly in a subsequent round after rewriting.}
\label{tab: rewritten and corrected UND}
\end{table*}

\end{document}